\documentclass[runningheads]{llncs}
\usepackage[T1]{fontenc}
\usepackage{graphicx}
\usepackage{booktabs}
\usepackage[misc]{ifsym}
\newcommand{\corr}{(\Letter)}

\usepackage{multirow}
\usepackage{array}
\usepackage{subcaption}
\usepackage{amsmath}
\usepackage{amssymb}
\usepackage{url}
\usepackage{adjustbox}
\usepackage{microtype}
\usepackage[unicode,
    colorlinks=true,
    linkcolor=blue,
    urlcolor=blue,
    citecolor=blue,
    bookmarks=true,
    bookmarksnumbered=true]{hyperref}

\begin{document}

\title{Why Do Class-Dependent Evaluation Effects Occur with Time Series Feature Attributions? A Synthetic Data Investigation}
\titlerunning{Why Do Class-Dependent Evaluation Effects Occur?}
\authorrunning{G. Baer et al.}

\author{Gregor Baer\inst{1, 2}\orcidID{0009-0002-9918-1376} \corr{} \and
Isel Grau\inst{1, 2}\orcidID{0000-0002-8035-2887} \and
Chao Zhang\inst{2, 3}\orcidID{0000-0001-9811-1881} \and
Pieter Van Gorp \inst{1,2}\orcidID{0000-0001-5197-3986}}
\institute{Information Systems, Eindhoven University of Technology, Eindhoven, The Netherlands \\
\email{g.baer@tue.nl} \and 
Eindhoven Artificial Intelligence Systems Institute, Eindhoven University of Technology, Eindhoven, The Netherlands \and Human-Technology Interaction, Eindhoven University of Technology, Eindhoven, The Netherlands
}

\maketitle
\begin{abstract}
Evaluating feature attribution methods represents a critical challenge in explainable AI (XAI), as researchers typically rely on perturbation-based metrics when ground truth is unavailable. 
However, recent work reveals that these evaluation metrics can show different performance across predicted classes within the same dataset. These ``class-dependent evaluation effects'' raise questions about whether perturbation analysis reliably measures attribution quality, with direct implications for XAI method development and evaluation trustworthiness.
We investigate under which conditions these class-dependent effects arise by conducting controlled experiments with synthetic time series data where ground truth feature locations are known. We systematically vary feature types and class contrasts across binary classification tasks, then compare perturbation-based degradation scores with ground truth-based precision-recall metrics using multiple attribution methods. Our experiments demonstrate that class-dependent effects emerge with both evaluation approaches, even in simple scenarios with temporally localized features, triggered by basic variations in feature amplitude or temporal extent between classes.
Most critically, we find that perturbation-based and ground truth metrics frequently yield contradictory assessments of attribution quality across classes, with weak correlations between evaluation approaches. 
These findings suggest that researchers should interpret perturbation-based metrics with care, as they may not always align with whether attributions correctly identify discriminating features. 
By showing this disconnect, our work points toward reconsidering what attribution evaluation actually measures and developing more rigorous evaluation methods that capture multiple dimensions of attribution quality.
.
\end{abstract}

\section{Introduction}\label{h:intro}
As machine learning models of increasing complexity advance prediction accuracy across domains, including time series classification~\cite{middlehurst.etal_2024_bake}, the field of Explainable Artificial Intelligence (XAI) has emerged to answer why models make certain predictions. However, a fundamental challenge in XAI research continues to be the evaluation of explanations. It remains challenging to determine what makes an explanation ``good'' and how to best evaluate explanation quality without ground truth labels, unlike prediction tasks where true outcomes enable direct performance measurement.

While human-centered evaluation through user studies provides direct assessment of explanation utility~\cite{rong.etal_2024_humancentered}, such approaches demand substantial time and resources, limiting their use during XAI method development. Consequently, XAI researchers frequently employ functional evaluation techniques that computationally verify whether explanations satisfy desirable properties~\cite{doshi-velez.kim_2018_considerations,nauta.etal_2023_anecdotal}. This computational approach enables quicker iteration and systematic comparison across methods, making it essential for advancing XAI research.

Researchers have adapted feature attribution methods originally developed for images to time series classification and evaluated them functionally, including~\cite{schlegel.etal_2019_rigorous,simic.etal_2022_perturbation,schlegel.keim_2023_deep,turbe.etal_2023_evaluation,nguyen.etal_2024_robust,serramazza.etal_2024_improving}. 
These evaluations predominantly rely on perturbation analysis, first introduced by Samek et al.~\cite{samek.etal_2017_evaluating}, which operates under the assumption that modifying features identified as important should proportionally degrade model output. This approach has become the de facto standard for attribution evaluation in the absence of ground truth.

However, Baer et al.~\cite{baer.etal_2025_classdependent} recently identified a potential methodological limitation: ``class-dependent perturbation effects'', where perturbation-based metrics demonstrate different performance across predicted classes within datasets. This phenomenon creates a problem because benchmarks relying on average scores can mask class-specific biases, potentially leading researchers to incorrect conclusions about the quality of attribution methods. Although previous research acknowledges that perturbation results vary by perturbation strategy and recommends testing multiple approaches~\cite{simic.etal_2022_perturbation,schlegel.keim_2023_deep}, class-dependent effects seem to persist across various perturbation methods and datasets~\cite{baer.etal_2025_classdependent}. The conditions generating these effects remain unexplored, leaving a gap in our understanding of how to reliably evaluate feature attributions.

Understanding why these effects arise carries critical implications for XAI evaluation methodology. If class-dependent differences reflect genuine variations in attribution quality, where methods truly perform better for certain classes, then perturbation analysis correctly captures these differences. However, if perturbation analysis itself introduces systematic biases regardless of attribution quality, researchers may draw misleading conclusions about method performance. This distinction fundamentally affects how we interpret and trust XAI evaluation results.

Our research addresses this gap by investigating: \textbf{Under which conditions do class-dependent evaluation effects arise?} We approach this question through systematically constructed synthetic data experiments, creating binary time series classification tasks with known ground truth features localized in time. By varying feature types (level shifts, pulses, sine waves, trends) and class contrasts (amplitude, length), we isolate factors contributing to evaluation disparities.

This paper offers three primary contributions to XAI evaluation methodology. 
First, we identify specific conditions under which class-dependent evaluation differences emerge, demonstrating that these effects manifest even in minimally complex classification tasks. We show that simple variations in feature prevalence and amplitude across classes trigger class-dependent effects, even when the underlying discriminating feature remains structurally identical between classes.
Second, we show that class-dependent effects occur in both ground truth-based and perturbation-based evaluation metrics, indicating that attribution methods genuinely perform differently across classes rather than reflecting measurement artifacts.
Third, and most critically, we demonstrate that ground truth and perturbation metrics can yield contradictory assessments of attribution quality across classes, revealing that these approaches may measure fundamentally different aspects of attribution behavior. While ground truth metrics assess whether attributions locate discriminating features, perturbation analysis may capture other properties that do not necessarily align with feature identification accuracy.
By showing when and why evaluation metrics diverge, our work advances efforts toward more rigorous assessment frameworks in time series analysis as advocated by Theissler et al.~\cite{theissler.etal_2022_explainable}.

The remainder of this paper is organized as follows. Section~\ref{h:eval-framework} establishes our evaluation framework, including perturbation-based and ground truth-based metrics with synthetic data generation. Section~\ref{h:exp-design} details our experimental design across different feature types, class contrasts, and attribution methods. Section~\ref{h:results-discussion} presents findings on when class-dependent evaluation effects emerge and analyzes the correspondence between evaluation approaches. Section~\ref{h:conclusion} summarizes our contributions and implications for XAI evaluation methodology.

\section{Evaluation Framework}\label{h:eval-framework}

Feature attribution methods for time series classification identify the time points that influence a model's predictions. Since ground truth is typically unavailable for real-world data, perturbation analysis is commonly used to assess attribution quality by modifying input features and observing the impact on model predictions. The underlying assumption is that perturbing important time points should cause significant prediction changes, while modifying irrelevant time points should have minimal impact. In this work, we evaluate feature attributions using both perturbation-based metrics and ground truth comparison on synthetic data where true feature locations are known. 

\begin{figure}[!htbp]
\includegraphics[width=\textwidth]{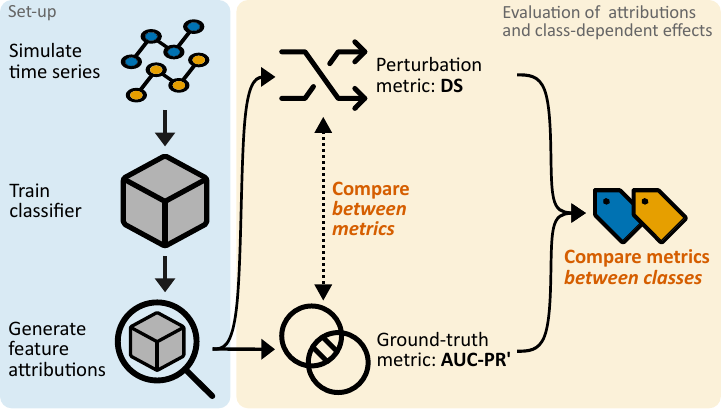}
\caption{Framework for analyzing class-dependent evaluation effects. We generate synthetic time series with features localized in time, train classifiers, and compare perturbation-based versus ground truth-based attribution evaluations to identify when class-dependent evaluation effects arise.}
\label{fig:cdpe-why-schematic}
\end{figure}

Figure~\ref{fig:cdpe-why-schematic} illustrates our evaluation framework for systematically investigating class-dependent effects. We generate synthetic time series with known discriminative features, compute attributions using multiple methods, and assess attribution quality through both perturbation-based and ground truth-based evaluations.

\noindent Table~\ref{tab:metrics-overview} outlines these complementary approaches, which capture different aspects of attribution quality. Perturbation analysis (Section~\ref{h:pert-metrics}) tests whether models rely on attributed regions for predictions, while ground truth evaluation (Section~\ref{h:gt-metrics}) verifies whether attributions correctly locate discriminative features. Although these metrics measure distinct properties, we would expect them to agree: attributions that correctly identify discriminative features should also prove critical when perturbed. Our experiments examine when and why these evaluation approaches converge or diverge across classes, showing whether class-dependent patterns reflect genuine attribution differences or evaluation inconsistencies.

\begin{table}[!htbp]
\caption{Attribution evaluation approaches employed in this study.}
\label{tab:metrics-overview}
\centering
\begin{tabular*}{\textwidth}{@{\extracolsep{\fill}} llp{8.5cm} @{}}
\toprule
Approach & Metric & Core question and method \\
\midrule
Perturbation & DS & Do predictions degrade when removing important features? \newline
\textit{Method:} Compare model confidence after perturbing most relevant vs. least relevant features. \\
Ground truth & AUC-PR$'$ & Do high attribution scores overlap with known discriminative feature regions? \newline
\textit{Method:} Measure how well attributions retrieve true feature locations using normalized precision-recall analysis. \\
\bottomrule
\bottomrule
\end{tabular*}
\end{table}

\subsection{Perturbation-Based Attribution Evaluation}\label{h:pert-metrics}

We use the degradation score (DS) for perturbation analysis, following the approach of Baer et al~\cite{baer.etal_2025_classdependent}. The DS, originally introduced by Schulz et al.~\cite{schulz.etal_2020_restricting} and first applied to time series by Šimić et al.~\cite{simic.etal_2022_perturbation}, compares attribution quality by contrasting two perturbation sequences: most relevant features first (MoRF) versus least relevant features first (LeRF).

Consider a univariate time series $\mathbf{x} = [x_1, \dots, x_N]$ where a trained classifier $f(\mathbf{x})$ generates class probability distributions with $q_c$ representing the confidence for class $c$. We evaluate feature attribution methods that produce temporal importance scores $\mathbf{r} = [r_1, \dots, r_N]$, where higher values of $r_i$ indicate greater relevance of timestep $i$ for the classifier's decision.

We perturb $\mathbf{x}$ by replacing timestep values with zero ($x_i'=0$) or Gaussian noise ($x'_i \sim \mathcal{N}(0, 1)$), creating two perturbation sequences: MoRF perturbs timesteps in descending order of attribution scores, while LeRF perturbs in ascending order. The DS compares these sequences:
\begin{equation}
    \text{DS} = \frac{1}{m} \sum_i^m (\text{PC}_{\text{LeRF}_i} - \text{PC}_{\text{MoRF}_i}),
\end{equation}
where $\text{PC}_{\text{order}_i}$ represents the predicted probabilities after perturbing $i$ features in the specified order. The DS ranges from -1 to 1, with positive values indicating effective attributions, i.e. MoRF perturbations degrade predictions more than LeRF.

\subsection{Synthetic Data Generation}\label{h:synth-data}

In addition to perturbation analysis, we construct synthetic time series data with known ground truth feature locations to directly evaluate attribution correctness. This allows us to assess whether attribution methods can correctly identify the specific time points where class-discriminating features are located.

Each synthetic time series $\mathbf{x}$ contains a single class-discriminating, contiguous feature over multiple time points embedded within a foundation signal through additive composition:
\begin{equation}
\mathbf{x} = \mathbf{n} + \mathbf{f},
\end{equation}
where $\mathbf{n} \sim \mathcal{N}(0, 1)$ represents background noise and $\mathbf{f}$ contains the class-specific temporal pattern within a designated feature window, with zeros elsewhere. Both vectors have identical length to enable element-wise addition.

\begin{figure}[!htbp]
\includegraphics[width=\textwidth]{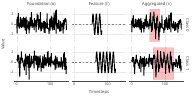}
\caption{Synthetic data generation example demonstrating feature length contrast. Two instances show identical sine wave features (red regions) with different temporal extents: Class 0 spans 20\% of timesteps while Class 1 spans 40\%.}
\label{fig:data-generation-example}
\end{figure}

While real-world time series features may exhibit complex temporal dependencies, we deliberately restrict investigation to temporally localized features occupying contiguous windows of predetermined length $L$. This design choice isolates evaluation effects under minimal complexity conditions, enabling attribution of observed phenomena to specific data characteristics rather than confounding factors.
Class-dependent variations are introduced through two systematic contrast mechanisms. \emph{Feature length contrast} varies temporal extent between classes, testing whether feature prevalence affects evaluation metrics. \emph{Amplitude contrast} varies signal magnitude between classes, examining whether signal strength differences drive evaluation disparities. Feature locations are randomized across observations to prevent positional classification shortcuts. For an example of this approach, see Figure~\ref{fig:data-generation-example}, which shows two possible instances with different classes from a simulated dataset. 

\subsection{Ground Truth-Based Attribution Evaluation}\label{h:gt-metrics}

To evaluate feature attributions on synthetic data, we treat attribution evaluation as a retrieval problem: can high attribution scores identify true feature locations? Let $\mathbf{l} = [l_1, \ldots, l_N]$ where $l_i \in \{0,1\}$ represent a binary mask indicating ground truth feature locations, with $l_i=1$ denoting the presence of a class-discriminating feature $f_i$ at time point $i$.
For attribution scores $\mathbf{r}$, we threshold at different values $t$ to obtain binary predictions $\hat{l}_i^{(t)} = I(r_i \geq t)$ of important feature locations. We use all unique attribution values as thresholds, evaluated in descending order, which ensures we capture all possible precision-recall operating points. At each threshold, we compute:
\begin{equation}
    \text{Precision}(t) = \frac{TP(t)}{TP(t) + FP(t)}, \quad \text{Recall}(t) = \frac{TP(t)}{P},
\end{equation}
where $TP(t) = \sum_{i=1}^{N} \hat{l}_i^{(t)} \cdot l_i$ (correctly identified feature locations), $FP(t) = \sum_{i=1}^{N} \hat{l}_i^{(t)} \cdot (1-l_i)$ (false feature locations), and $P = \sum_{i=1}^{N} l_i$ (total ground truth features).

To assess whether feature attributions correctly highlight crucial regions within the time series, we compute the AUC-PR. This metric calculates the area beneath the precision-recall curve, where precision is plotted as a function of recall:
\begin{equation}
    \text{AUC-PR} = \int_{0}^{1} \text{Precision}(\text{recall}) \, d(\text{recall}),
\end{equation}
computed using the trapezoidal rule over unique recall values. We choose AUC-PR because it suits our sparse ground truth features well. For binary classification problems, the random baseline equals the feature prevalence, $\text{Prevalence}=\frac{P}{N}$.

Since we vary feature lengths across experiments, we normalize the AUC-PR to enable fair comparison across different prevalence rates:
\begin{equation}
    \text{AUC-PR}' = \frac{\text{AUC-PR} - \text{Prevalence}}{1 - \text{Prevalence}}.
\end{equation}
This normalized score represents improvement over the random baseline, with values approaching 1 indicating perfect performance and negative values indicating worse-than-random attribution quality.

\section{Experimental Design}\label{h:exp-design}

\textbf{Dataset specifications.} 
We construct synthetic datasets consisting of univariate time series with 150 timesteps for binary classification tasks. Each dataset comprises 1000 training samples, 300 validation samples, and 300 test samples with balanced class distributions. While all sets are generated from the same data-generating process, we use different random seeds to ensure independence. Class-discriminating features are randomly positioned within each time series, with feature windows $[t_{\text{start}}, t_{\text{start}} + L - 1]$ varying across observations to prevent positional classification shortcuts. All instances undergo standardization prior to model training.

\textbf{Feature types.}
We implement four feature types to represent common temporal patterns: (1) level shifts representing constant amplitude changes, (2) Gaussian pulses capturing transient spikes, (3) sine waves modeling periodic oscillations reminiscent of seasonality, and (4) local trends reflecting gradual directional changes. Definitions and parameters for each feature type are provided in Table~\ref{tab:feature-definitions}.

\begin{table}[!htbp]
\centering
\caption{Feature type definitions. Formulas show signal generation within the feature window, where $t \in [0, L-1]$ is the relative timestep. Parameters: $A=$ amplitude, $L=$ feature window length, $p=$ period, $\sigma = L/6=$ pulse width.}
\label{tab:feature-definitions}
\begin{tabular*}{\textwidth}{@{\extracolsep{\fill}}l c l l}
\toprule
Feature & Shape & Description & Definition \\
\midrule
Level & \includegraphics[width=1.1cm,height=0.74cm]{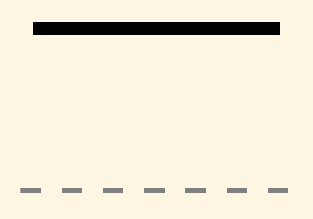} & Constant amplitude/level shift & $f(t) = A$ \\
Pulse & \includegraphics[width=1.1cm,height=0.74cm]{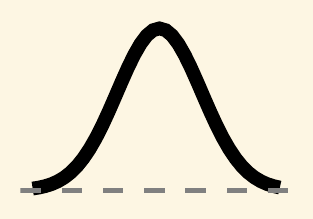} & Gaussian-shaped transient signal & $f(t) = A e^{-\frac{(t - L/2)^2}{2\sigma^2}}, \quad \sigma = L/6$ \\
Sine & \includegraphics[width=1.1cm,height=0.74cm]{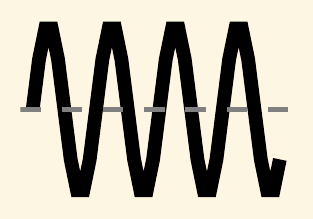} & Periodic oscillation & $f(t) = A \sin(2\pi t/p), \quad p = 10$ \\
Trend & \includegraphics[width=1.1cm,height=0.74cm]{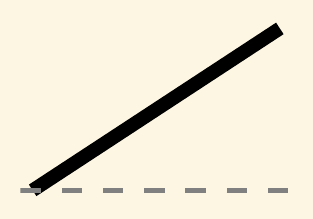} & Linear increase from zero & $f(t) = At/(L-1)$ \\
\bottomrule
\end{tabular*}
\end{table}

\begin{table}[!htbp]
\centering
\caption{Experimental class contrasts applied to all four feature types (Level, Pulse, Sine, Trend; see Table~\ref{tab:feature-definitions}). Parameters: $A=$ feature amplitude, $L=$ feature window length (timesteps), $\mathbf{n} =$ background noise. }
\label{tab:contrast-types}
\begin{tabular*}{\textwidth}{@{\extracolsep{\fill}}lllll@{}}
\toprule
Contrast & Class 0 & Class 1 & Both classes & Purpose \\
\midrule
Amplitude & $A = 1$ & $A = 2$ & $\mathbf{n} \sim \mathcal{N}(0, 1)$, $L=60$  & Test magnitude effects \\
Length    & $L=30$  & $L=60$  & $\mathbf{n} \sim \mathcal{N}(0, 1)$, $A=2$  & Test prevalence effects \\
\bottomrule
\end{tabular*}
\end{table}

\textbf{Class contrasts.}
We create binary classification tasks by systematically contrasting features across two dimensions: amplitude differences to examine magnitude-based discrimination, and length differences to investigate temporal prevalence effects. 
This design isolates how different types of class distinctions influence attribution evaluation metrics.

We embed features within background Gaussian noise, with feature amplitudes chosen to maintain detectability while avoiding trivial separability. Feature lengths span 20\% to 40\% of the time series to represent moderate temporal prevalence. Combining four feature types with two contrast mechanisms yields eight datasets. 
Table~\ref{tab:contrast-types} summarizes the specific class differences for each contrast type. For example, amplitude-contrast datasets vary only feature magnitude while maintaining identical temporal extent, whereas length-contrast datasets vary temporal coverage while preserving signal strength.

\textbf{Model architecture and training.}
We employ two established deep learning architectures for time series classification: ResNet~\cite{wang.etal_2017_time} and InceptionTime~\cite{ismailfawaz.etal_2020_inceptiontime}, both of which offer robust performance across time series benchmarks~\cite{fawaz.etal_2019_deep}.
We train models using AdamW optimization with cosine annealing learning rate scheduling and early stopping (patience = 10 epochs) based on validation performance over a maximum of 100 epochs.
Both architectures achieve test accuracy and weighted F1 scores exceeding 0.9 across all datasets, with one exception: InceptionTime on the Amplitude-Pulse dataset (F1 = 0.7). 
These performance levels ensure that observed class-dependent evaluation differences stem from attribution or perturbation methodology rather than classifier bias toward specific classes. 
Complete performance metrics are provided in Table~\ref{tab:model_performance} in Appendix~\ref{appendix}.

\textbf{Attribution methods.}
We evaluate three attribution methods: Gradients (GR)~\cite{simonyan.etal_2014_deep}, Integrated Gradients (IG)~\cite{sundararajan.etal_2017_axiomatic}, and Feature Occlusion (FO)~\cite{fong.vedaldi_2017_interpretable}. IG and FO were selected because they demonstrated both high attribution quality and pronounced class-dependent evaluation effects in previous research~\cite{baer.etal_2025_classdependent}, making them suitable candidates for investigating the conditions under which such effects arise. We include GR as a computationally efficient baseline to ensure our findings span different levels of attribution complexity. The selection covers both gradient-based approaches (GR and IG) and perturbation-based attribution (FO). We compute attributions for all 300 test samples, explaining each model's predicted class label.

\textbf{Perturbation analysis.}
We employ two perturbation strategies: replacing values with zero and substituting with Gaussian noise. Both strategies are commonly used in time series attribution evaluation~\cite{schlegel.keim_2023_deep,simic.etal_2022_perturbation,turbe.etal_2023_evaluation}.
Since we standardize all time series, zero substitution effectively replaces values with the sample mean. For Gaussian noise perturbation, we sample from the same distribution as the background noise ($\mathcal{N}(0,1)$), following the controlled masking approach used in existing benchmarks~\cite{ismail_benchmarking_2020}. This approach ensures that perturbations align more closely with the data distribution, avoiding the potential artifacts of more extreme substitution methods or constant replacement values.

We implement perturbations according to MoRF and LeRF orderings determined by attribution scores, substituting features incrementally one timestep at a time. To maintain fair comparisons across datasets with varying feature lengths, we perturb exactly $L$ timesteps for each sample, where $L$ corresponds to the true feature window size. For instance, in length-contrast datasets, we perturb 30 timesteps for class 0 samples and 60 timesteps for class 1 samples. We record predicted probabilities after each perturbation step to calculate the DS.

\section{Results and Discussion}\label{h:results-discussion}

\begin{figure}[!htbp]
\includegraphics[width=\textwidth]{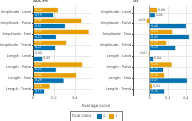}
\caption{Average $\text{AUC-PR}'$ (left) and DS (right) by dataset and true class. $\text{AUC-PR}'$ measures attribution overlap with known features; DS compares prediction degradation when perturbing regions with high versus low attribution scores. Both metrics show class-dependent differences across datasets, with opposing performance patterns between classes in most cases.}\label{fig:avg-metrics}
\end{figure}

Figure~\ref{fig:avg-metrics} presents the $\text{AUC-PR}'$ and DS across datasets and true class labels (detailed results for all experimental conditions appear in Appendix~\ref{appendix}). 
We aggregate results across attribution methods and model architectures to emphasize general patterns in class-dependent evaluation effects, though individual method-specific variations exist within these trends.

\textbf{Ground truth-based evaluation.}
The ground truth evaluation reveals substantial variation in how well attribution methods recover true class-discriminating feature regions, both across datasets and between classes within datasets. Class 1 observations (with higher amplitude or longer feature windows) consistently achieve higher $\text{AUC-PR}'$ scores than class 0 observations across seven of eight datasets. The Length-Level dataset is the sole exception where class 1 attributions approach random performance. 
Overall, Pulse, Sine, and Trend features demonstrate pronounced class-dependent differences in attribution quality. The disparities prove particularly large for Amplitude-Sine and Length-Pulse datasets: class 1 observations in both achieve $\text{AUC-PR}'$ scores more than double those of class 0 observations, indicating that attribution methods recover ground truth features more effectively for one class than the other.

\textbf{Perturbation-based evaluation.}
Perturbation-based evaluation through DS scores reveals patterns that directly contradict the ground truth findings. While ground truth evaluation favors class 1 attributions in seven of eight datasets, perturbation analysis indicates better attribution quality for class 0 observations in six of eight datasets. This systematic reversal across most feature types and contrast mechanisms suggests the two evaluation approaches measure different aspects of attribution behavior.
Length-Level and Amplitude-Level datasets represent exceptions where both evaluation approaches converge. These datasets show smaller class discrepancies and assign relatively low scores across both metrics. This pattern likely reflects how attribution methods focus on temporal change points rather than extended feature windows when detecting level shifts. Such localized attribution behavior reduces overlap with our ground truth evaluation windows and limits perturbation impact when removing individual timesteps.

Overall, we observe an inverse relationship: datasets where one class achieves the highest attribution performance also tend to show the strongest disagreement between evaluation approaches. This pattern appears consistently, revealing tensions between ground truth and perturbation-based assessment even when individual metrics indicate strong performance.

\noindent \textbf{Correspondence between evaluation approaches.}
To examine the relationship between evaluation approaches, we compute Spearman correlations between $\text{AUC-PR}'$ and DS scores. We use Spearman correlation because the metrics operate on different scales and we focus on monotonic relationships, that is, whether higher scores in one metric correspond to higher scores in the other.

\begin{figure}[!htbp]
\includegraphics[width=\textwidth]{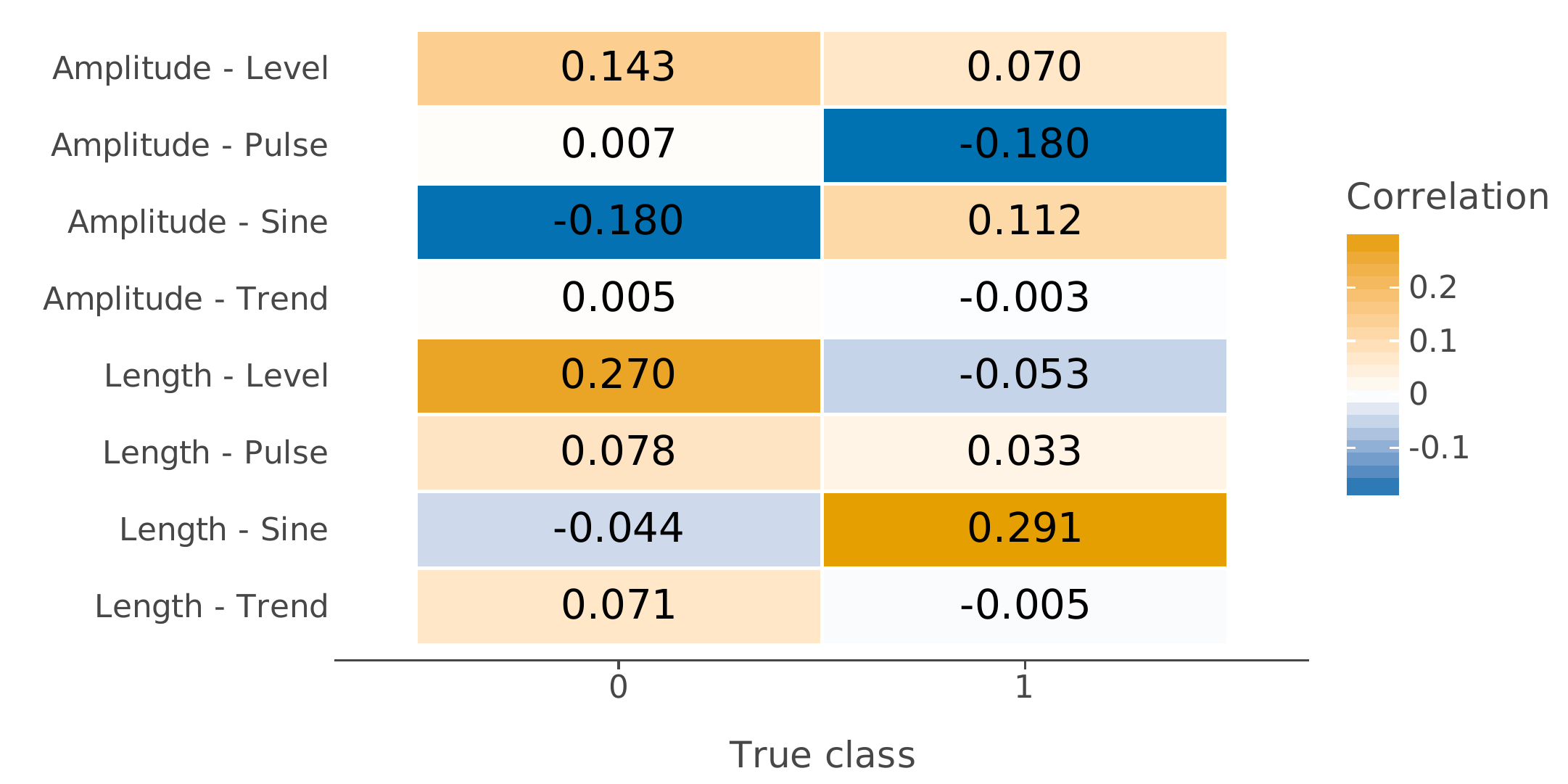}
\caption{Average Spearman correlation between $\text{AUC-PR}'$ and DS by dataset and true class. Correlations are computed per observation, then averaged. Weak correlations and sign reversals between classes show that ground truth and perturbation-based metrics can disagree about attribution quality.}\label{fig:metric-correlations}
\end{figure}

Figure~\ref{fig:metric-correlations} shows correlations per dataset and class. The analysis confirms our earlier observations about contradictory class-dependent evaluation effects. Correlations remain weak to moderate, ranging from -0.18 to 0.291, and vary substantially between classes within datasets. For example, Amplitude-Sine shows a negative correlation (-0.18) for class 0 but a positive correlation (0.112) for class 1. Length-Sine achieves the highest correlation (0.291) for class 1 while class 0 shows a slight negative correlation.

\noindent \textbf{Summary.} 
Class-dependent evaluation differences emerge across almost all feature types and contrast mechanisms. Ground truth and perturbation-based metrics yield contradictory assessments: while ground truth evaluation favors class 1 attributions in seven of eight datasets, perturbation analysis indicates better performance for class 0 in six of eight cases. Spearman correlations between evaluation approaches remain weak and inconsistent, ranging from -0.18 to 0.291 across datasets and classes. This weak correspondence demonstrates that perturbation analysis does not reliably identify whether attributions correctly locate class-discriminating regions, indicating that the two evaluation frameworks measure different aspects of attribution behavior.

\section{Conclusion}\label{h:conclusion}

We investigate conditions under which class-dependent evaluation effects arise in time series feature attribution assessment. Through controlled synthetic experiments with known ground truth features, we compare perturbation-based and ground truth-based evaluation metrics across multiple attribution methods.
Our results demonstrate that even simple variations in feature characteristics, such as amplitude differences or temporal extent, can trigger pronounced evaluation disparities between classes. These class-dependent effects emerge consistently across feature types and contrast mechanisms. Most notably, perturbation-based metrics frequently suggest superior attribution quality for one class while ground truth metrics indicate the opposite.

This systematic disagreement creates a conceptual challenge: in the absence of ground truth, researchers rely on perturbation-based metrics knowing they measure model sensitivity, with the implicit assumption that this serves as a reasonable proxy for feature localization accuracy. Our findings challenge this assumption by demonstrating that model sensitivity and correct feature identification can diverge systematically. The disagreement is particularly concerning since real-world datasets lack ground truth references, leaving perturbation as the primary evaluation option. Researchers should interpret perturbation metrics cautiously, recognizing they may not accurately reflect whether attributions succeed in identifying discriminative feature locations despite our expectations that these properties should align.

Our findings show such disagreements emerge even under minimal complexity. However, several limitations constrain the generalizability of our findings.
We examine only simple univariate time series with temporally localized features, employing two deep learning architectures and three general-purpose attribution methods. While this design enables controlled analysis, it excludes complex temporal patterns and specialized attribution approaches designed for time series. 
Our evaluation also assumes that good attributions should cover entire feature windows, yet models may legitimately focus on sparse temporal markers like change points or learn from subtle patterns beyond our designed features. When models identify features through different strategies than we expect, standard metrics may underestimate their actual attribution quality. This mismatch between how we measure attribution performance and how models process temporal information highlights the challenge of defining ``correct'' attributions in time series contexts.

Future research can address both empirical and conceptual dimensions of these evaluation challenges. Testing time series-specific attribution methods like LimeSegment~\cite{sivill.flach_2022_limesegment} may reveal whether specialized approaches reduce class-dependent effects. Expanding synthetic datasets to include multiple discriminating patterns and more complex temporal dependencies would test generalizability beyond our localized feature settings. The weak correspondence between perturbation and ground truth metrics suggests that measuring model sensitivity may not adequately assess feature localization accuracy. This disconnect between what we aim to measure and what perturbation analysis actually captures encourages developing more rigorous evaluation frameworks that acknowledge different dimensions of attribution quality.

\clearpage
\appendix
\section{Detailed Experimental Results} \label{appendix} 
\begin{table}[!htbp]
\caption{Model performance by dataset on the test set.}
\label{tab:model_performance}
\centering
\setlength{\tabcolsep}{4pt}
\begin{tabular*}{0.8\textwidth}{@{\extracolsep{\fill}} llcccc @{}}
\toprule
 &  & \multicolumn{2}{c}{ResNet} & \multicolumn{2}{c}{InceptionTime} \\
\cmidrule(lr){3-4} \cmidrule(lr){5-6}
Contrast & Feature & Accuracy & F1 & Accuracy & F1 \\
\midrule
\multirow{4}{*}{Amplitude} & Level & 0.990 & 0.990 & 0.980 & 0.980 \\
 & Pulse & 0.900 & 0.899 & 0.723 & 0.700 \\
 & Sine & 0.997 & 0.997 & 1.000 & 1.000 \\
 & Trend & 0.920 & 0.920 & 0.940 & 0.940 \\
\cmidrule{1-6}
\multirow{4}{*}{Length} & Level & 1.000 & 1.000 & 1.000 & 1.000 \\
 & Pulse & 0.923 & 0.923 & 0.947 & 0.947 \\
 & Sine & 0.997 & 0.997 & 1.000 & 1.000 \\
 & Trend & 0.927 & 0.927 & 0.973 & 0.973 \\
\bottomrule
\end{tabular*}
\end{table}

\begin{table}[!htbp]
\caption{Average $\text{AUC-PR}'$ by classifier, dataset, attribution method and true class.}
\label{tab:auc_pr}
\centering
\setlength{\tabcolsep}{4pt}
\begin{tabular*}{\textwidth}{@{\extracolsep{\fill}} lllcccccc @{}}
\toprule
 & & & \multicolumn{2}{c}{GR} & \multicolumn{2}{c}{IG} & \multicolumn{2}{c}{FO} \\
\cmidrule(lr){4-5} \cmidrule(lr){6-7} \cmidrule(lr){8-9}
Model & Contrast & Feature & C0 & C1 & C0 & C1 & C0 & C1 \\
\midrule
\multirow{8}{*}{ResNet} & \multirow{4}{*}{Amplitude} & Level & .134 & .234 & .435 & .384 & -.032 & -.068 \\
  &  & Pulse & .235 & .313 & .188 & .284 & .151 & .270 \\
  &  & Sine & .219 & .357 & .170 & .262 & .308 & .681 \\
  &  & Trend & .126 & .127 & .074 & .278 & .058 & .058 \\
\cmidrule{2-9}
  & \multirow{4}{*}{Length} & Level & .190 & .483 & .038 & -.101 & .102 & -.038 \\
  &  & Pulse & .179 & .329 & .226 & .313 & .158 & .197 \\
  &  & Sine & .262 & .358 & .139 & .166 & .304 & .455 \\
  &  & Trend & .004 & .129 & -.018 & -.024 & -.004 & -.013 \\
\midrule
\multirow{8}{*}{InceptionTime} & \multirow{4}{*}{Amplitude} & Level & .280 & .503 & .275 & .225 & .071 & .148 \\
  &  & Pulse & .574 & .750 & .264 & .572 & .427 & .599 \\
  &  & Sine & .139 & .573 & .256 & .645 & .219 & .681 \\
  &  & Trend & .345 & .528 & .451 & .588 & .223 & .348 \\
\cmidrule{2-9}
  & \multirow{4}{*}{Length} & Level & .033 & -.002 & .147 & -.101 & .025 & -.183 \\
  &  & Pulse & .274 & .733 & .248 & .711 & .226 & .555 \\
  &  & Sine & .226 & .300 & .512 & .684 & .312 & .528 \\
  &  & Trend & .285 & .386 & .208 & .306 & .158 & .166 \\
\bottomrule
\end{tabular*}
\end{table}

\begin{table}[!htbp]
\caption{Average DS by classifier, dataset, perturbation strategy, attribution method, and true class.}
\label{tab:ds}
\centering
\small
\setlength{\tabcolsep}{2pt}
\begin{tabular*}{\textwidth}{@{\extracolsep{\fill}} llllcccccc @{}}
\toprule
 & & & & \multicolumn{2}{c}{GR} & \multicolumn{2}{c}{IG} & \multicolumn{2}{c}{FO} \\
\cmidrule(lr){5-6} \cmidrule(lr){7-8} \cmidrule(lr){9-10}
Model & Contrast & Feature & Perturbation & C0 & C1 & C0 & C1 & C0 & C1 \\
\midrule
\multirow{16}{*}{ResNet} & \multirow{8}{*}{Amplitude} & \multirow{2}{*}{Level} & Gaussian & .133 & -.460 & .023 & .185 & -.001 & .054 \\
  &  &  & Zero & .155 & -.430 & -.027 & .508 & .013 & .043 \\
  &  & \multirow{2}{*}{Pulse} & Gaussian & .510 & -.296 & .543 & -.093 & .340 & .042 \\
  &  &  & Zero & .639 & -.220 & .712 & -.026 & .480 & .086 \\
  &  & \multirow{2}{*}{Sine} & Gaussian & .266 & .083 & .551 & -.072 & .334 & .229 \\
  &  &  & Zero & .511 & .043 & .778 & -.067 & .561 & .231 \\
  &  & \multirow{2}{*}{Trend} & Gaussian & .474 & -.204 & .391 & -.020 & .286 & .057 \\
  &  &  & Zero & .634 & -.199 & .446 & .051 & .336 & .085 \\
\cmidrule{2-10}
  & \multirow{8}{*}{Length} & \multirow{2}{*}{Level} & Gaussian & .116 & -.142 & .010 & -.071 & -.029 & .032 \\
  &  &  & Zero & .090 & -.140 & .016 & .081 & .004 & .171 \\
  &  & \multirow{2}{*}{Pulse} & Gaussian & .314 & -.179 & .146 & .086 & .150 & .062 \\
  &  &  & Zero & .283 & -.140 & .078 & .353 & .111 & .246 \\
  &  & \multirow{2}{*}{Sine} & Gaussian & .439 & -.066 & .505 & -.030 & .399 & .074 \\
  &  &  & Zero & .677 & -.023 & .737 & -.029 & .621 & .019 \\
  &  & \multirow{2}{*}{Trend} & Gaussian & .378 & -.291 & .116 & -.099 & .179 & -.069 \\
  &  &  & Zero & .347 & -.353 & -.020 & .204 & .034 & .158 \\
\midrule
\multirow{16}{*}{InceptionTime} & \multirow{8}{*}{Amplitude} & \multirow{2}{*}{Level} & Gaussian & .187 & -.157 & .120 & .114 & .049 & .187 \\
  &  &  & Zero & .089 & .049 & -.006 & .449 & .005 & .459 \\
  &  & \multirow{2}{*}{Pulse} & Gaussian & .390 & -.191 & .169 & -.095 & .263 & -.122 \\
  &  &  & Zero & .420 & .124 & .089 & .252 & .268 & .129 \\
  &  & \multirow{2}{*}{Sine} & Gaussian & .230 & .458 & .383 & .412 & .261 & .434 \\
  &  &  & Zero & .371 & .453 & .537 & .415 & .417 & .428 \\
  &  & \multirow{2}{*}{Trend} & Gaussian & .256 & .112 & .181 & .331 & .199 & .273 \\
  &  &  & Zero & .207 & .426 & -.011 & .718 & .027 & .627 \\
\cmidrule{2-10}
  & \multirow{8}{*}{Length} & \multirow{2}{*}{Level} & Gaussian & .032 & -.058 & .110 & -.078 & .021 & -.060 \\
  &  &  & Zero & .017 & -.116 & .115 & .038 & .009 & .164 \\
  &  & \multirow{2}{*}{Pulse} & Gaussian & .179 & .273 & .112 & .351 & .167 & .291 \\
  &  &  & Zero & .152 & .508 & .021 & .538 & .105 & .540 \\
  &  & \multirow{2}{*}{Sine} & Gaussian & .178 & .242 & .227 & .456 & .206 & .428 \\
  &  &  & Zero & .293 & .150 & .347 & .387 & .330 & .362 \\
  &  & \multirow{2}{*}{Trend} & Gaussian & .162 & -.087 & .259 & .135 & .210 & .101 \\
  &  &  & Zero & .093 & .021 & .066 & .404 & .145 & .245 \\
\bottomrule
\end{tabular*}
\end{table}
\clearpage

\begin{credits}
\subsubsection{\ackname} This paper is supported by the European Union’s HORIZON Research and Innovation Program under grant agreement No. 101120657, project ENFIELD (European Lighthouse to Manifest Trustworthy and Green AI).

\subsubsection{\discintname}
All authors declare that they have no conflicts of interest.
\end{credits}

\bibliographystyle{splncs04}
\bibliography{references}

\begin{thebibliography}{10}
\providecommand{\url}[1]{\texttt{#1}}
\providecommand{\urlprefix}{URL }
\providecommand{\doi}[1]{https://doi.org/#1}

\bibitem{baer.etal_2025_classdependent}
Baer, G., Grau, I., Zhang, C., Van~Gorp, P.: Class-dependent perturbation effects in evaluating time series attributions. In: Proceedings of the 3rd World Conference on {{eXplainable}} Artificial Intelligence ({{XAI-2025}}) (2025), in Press. Preprint available: arXiv:2502.17022

\bibitem{doshi-velez.kim_2018_considerations}
{Doshi-Velez}, F., Kim, B.: Considerations for {{Evaluation}} and {{Generalization}} in {{Interpretable Machine Learning}}. In: Escalante, H.J., Escalera, S., Guyon, I., Bar{\'o}, X., G{\"u}{\c c}l{\"u}t{\"u}rk, Y., G{\"u}{\c c}l{\"u}, U., {van Gerven}, M. (eds.) Explainable and {{Interpretable Models}} in {{Computer Vision}} and {{Machine Learning}}, pp. 3--17. Springer (2018). \doi{10.1007/978-3-319-98131-4_1}

\bibitem{fawaz.etal_2019_deep}
Fawaz, H.I., Forestier, G., Weber, J., Idoumghar, L., Muller, P.A.: Deep learning for time series classification: A review. Data Mining and Knowledge Discovery  \textbf{33},  917--963 (2019). \doi{10.1007/s10618-019-00619-1}

\bibitem{fong.vedaldi_2017_interpretable}
Fong, R.C., Vedaldi, A.: Interpretable {{Explanations}} of {{Black Boxes}} by {{Meaningful Perturbation}}. In: Proceedings of the {{IEEE International Conference}} on {{Computer Vision}}. pp. 3429--3437 (2017), \url{https://openaccess.thecvf.com/content_iccv_2017/html/Fong_Interpretable_Explanations_of_ICCV_2017_paper.html}

\bibitem{ismail_benchmarking_2020}
Ismail, A.A., Gunady, M., Corrada~Bravo, H., Feizi, S.: Benchmarking deep learning interpretability in time series predictions. Advances in neural information processing systems  \textbf{33},  6441--6452 (2020), \url{https://proceedings.neurips.cc/paper_files/paper/2020/hash/47a3893cc405396a5c30d91320572d6d-Abstract.html}

\bibitem{ismailfawaz.etal_2020_inceptiontime}
Ismail~Fawaz, H., Lucas, B., Forestier, G., Pelletier, C., Schmidt, D.F., Weber, J., Webb, G.I., Idoumghar, L., Muller, P.A., Petitjean, F.: {{InceptionTime}}: {{Finding AlexNet}} for time series classification. Data Mining and Knowledge Discovery  \textbf{34},  1936--1962 (2020). \doi{10.1007/s10618-020-00710-y}

\bibitem{middlehurst.etal_2024_bake}
Middlehurst, M., Sch{\"a}fer, P., Bagnall, A.: Bake off redux: A review and experimental evaluation of recent time series classification algorithms. Data Mining and Knowledge Discovery  \textbf{38},  1958--2031 (2024). \doi{10.1007/s10618-024-01022-1}

\bibitem{nauta.etal_2023_anecdotal}
Nauta, M., Trienes, J., Pathak, S., Nguyen, E., Peters, M., Schmitt, Y., Schl{\"o}tterer, J., Van~Keulen, M., Seifert, C.: From {{Anecdotal Evidence}} to {{Quantitative Evaluation Methods}}: {{A Systematic Review}} on {{Evaluating Explainable AI}}. ACM Computing Surveys  \textbf{55},  1--42 (2023). \doi{10.1145/3583558}

\bibitem{nguyen.etal_2024_robust}
Nguyen, T.T., Le~Nguyen, T., Ifrim, G.: Robust explainer recommendation for time series classification. Data Mining and Knowledge Discovery  \textbf{38},  3372--3413 (2024). \doi{10.1007/s10618-024-01045-8}

\bibitem{rong.etal_2024_humancentered}
Rong, Y., Leemann, T., Nguyen, T.T., Fiedler, L., Qian, P., Unhelkar, V., Seidel, T., Kasneci, G., Kasneci, E.: Towards {{Human-Centered Explainable AI}}: {{A Survey}} of {{User Studies}} for {{Model Explanations}}. IEEE Transactions on Pattern Analysis and Machine Intelligence  \textbf{46},  2104--2122 (2024). \doi{10.1109/TPAMI.2023.3331846}

\bibitem{samek.etal_2017_evaluating}
Samek, W., Binder, A., Montavon, G., Lapuschkin, S., M{\"u}ller, K.R.: Evaluating the {{Visualization}} of {{What}} a {{Deep Neural Network Has Learned}}. IEEE Transactions on Neural Networks and Learning Systems  \textbf{28},  2660--2673 (2017). \doi{10.1109/TNNLS.2016.2599820}

\bibitem{schlegel.etal_2019_rigorous}
Schlegel, U., Arnout, H., {El-Assady}, M., Oelke, D., Keim, D.A.: Towards a rigorous evaluation of {{XAI}} methods on time series. In: 2019 {{IEEE}}/{{CVF}} International Conference on Computer Vision Workshop ({{ICCVW}}). pp. 4197--4201 (2019). \doi{10.1109/ICCVW.2019.00516}

\bibitem{schlegel.keim_2023_deep}
Schlegel, U., Keim, D.A.: A {{Deep Dive}} into {{Perturbations}} as {{Evaluation Technique}} for {{Time Series XAI}}. In: Longo, L. (ed.) Explainable {{Artificial Intelligence}}, vol.~1903, pp. 165--180. Springer Nature Switzerland, Cham (2023). \doi{10.1007/978-3-031-44070-0_9}

\bibitem{schulz.etal_2020_restricting}
Schulz, K., Sixt, L., Tombari, F., Landgraf, T.: Restricting the flow: {{Information}} bottlenecks for attribution. In: International Conference on Learning Representations (2020), \url{https://openreview.net/forum?id=S1xWh1rYwB}

\bibitem{serramazza.etal_2024_improving}
Serramazza, D.I., Nguyen, T.L., Ifrim, G.: Improving the~{{Evaluation}} and~{{Actionability}} of~{{Explanation Methods}} for~{{Multivariate Time Series Classification}}. In: Bifet, A., Davis, J., Krilavi{\v c}ius, T., Kull, M., Ntoutsi, E., {\v Z}liobait{\.e}, I. (eds.) Machine {{Learning}} and {{Knowledge Discovery}} in {{Databases}}. {{Research Track}}. pp. 177--195 (2024). \doi{10.1007/978-3-031-70359-1_11}

\bibitem{simic.etal_2022_perturbation}
{\v S}imi{\'c}, I., Sabol, V., Veas, E.: Perturbation effect: A metric to counter misleading validation of feature attribution. In: Proceedings of the 31st {{ACM}} International Conference on Information \& Knowledge Management. pp. 1798--1807 (2022). \doi{10.1145/3511808.3557418}

\bibitem{simonyan.etal_2014_deep}
Simonyan, K., Vedaldi, A., Zisserman, A.: Deep inside convolutional networks: Visualising image classification models and saliency maps. In: Proceedings of the {{International Conference}} on {{Learning Representations}} ({{ICLR}}) (2014). \doi{10.48550/arXiv.1312.6034}

\bibitem{sivill.flach_2022_limesegment}
Sivill, T., Flach, P.: {{LIMESegment}}: {{Meaningful}}, {{Realistic Time Series Explanations}}. In: Proceedings of {{The}} 25th {{International Conference}} on {{Artificial Intelligence}} and {{Statistics}}. pp. 3418--3433 (2022), \url{https://proceedings.mlr.press/v151/sivill22a.html}

\bibitem{sundararajan.etal_2017_axiomatic}
Sundararajan, M., Taly, A., Yan, Q.: Axiomatic {{Attribution}} for {{Deep Networks}}. In: Proceedings of the 34th {{International Conference}} on {{Machine Learning}}. pp. 3319--3328 (2017), \url{https://proceedings.mlr.press/v70/sundararajan17a.html}

\bibitem{theissler.etal_2022_explainable}
Theissler, A., Spinnato, F., Schlegel, U., Guidotti, R.: Explainable {{AI}} for {{Time Series Classification}}: {{A Review}}, {{Taxonomy}} and {{Research Directions}}. IEEE Access  \textbf{10},  100700--100724 (2022). \doi{10.1109/ACCESS.2022.3207765}

\bibitem{turbe.etal_2023_evaluation}
Turb{\'e}, H., Bjelogrlic, M., Lovis, C., Mengaldo, G.: Evaluation of post-hoc interpretability methods in time-series classification. Nature Machine Intelligence  \textbf{5},  250--260 (2023). \doi{10.1038/s42256-023-00620-w}

\bibitem{wang.etal_2017_time}
Wang, Z., Yan, W., Oates, T.: Time series classification from scratch with deep neural networks: {{A}} strong baseline. In: 2017 {{International Joint Conference}} on {{Neural Networks}} ({{IJCNN}}). pp. 1578--1585 (2017). \doi{10.1109/IJCNN.2017.7966039}

\end{thebibliography}

\end{document}